\documentclass[runningheads]{llncs}

\usepackage{eccv}

\usepackage{eccvabbrv}

\usepackage{graphicx}
\usepackage{booktabs}

\usepackage[accsupp]{axessibility}  

\usepackage{hyperref}

\usepackage{orcidlink}

\begin{document}
\begin{sloppypar}
\title{S2DM: Sector-Shaped Diffusion Models for Video Generation} 


\author{Haoran Lang\inst{1} \and
Yuxuan Ge\inst{1} \and
Zheng Tian\inst{1}\thanks{Corresponding author. Email: tianzheng@shanghaitech.edu.cn}}

\authorrunning{Lang et al.}

\institute{ShanghaiTech University, Shanghai, China \\
\email{\{langhr2022, geyx2023, tianzheng\}@shanghaitech.edu.cn}}

\maketitle

\begin{abstract}
\label{abstract}
 Diffusion models have achieved great success in image generation. However, when leveraging this idea for video generation, we face significant challenges in maintaining the consistency and continuity across video frames. This is mainly caused by the lack of an effective framework to align frames of videos with desired temporal features while preserving consistent semantic and stochastic features. In this work, we propose a novel Sector-Shaped Diffusion Model (S2DM) whose sector-shaped diffusion region is formed by a set of ray-shaped reverse diffusion processes starting at the same noise point. S2DM can generate a group of intrinsically related data sharing the same semantic and stochastic features while varying on temporal features with appropriate guided conditions. We apply S2DM to video generation tasks, and explore the use of optical flow as temporal conditions. Our experimental results show that S2DM outperforms many existing methods in the task of video generation without any temporal-feature modelling modules. For text-to-video generation tasks where temporal conditions are not explicitly given, we propose a two-stage generation strategy which can decouple the generation of temporal features from semantic-content features. We show that, without additional training, our model integrated with another temporal conditions generative model can still achieve comparable performance with existing works. Our results can be viewd at~\href{https://s2dm.github.io/S2DM/}{https://s2dm.github.io/S2DM/}.
 \vspace{-7mm}
\end{abstract}
\begin{figure}[ht]
\vspace{-2mm}
\begin{center}
\centerline{\includegraphics[width=1.0\linewidth]{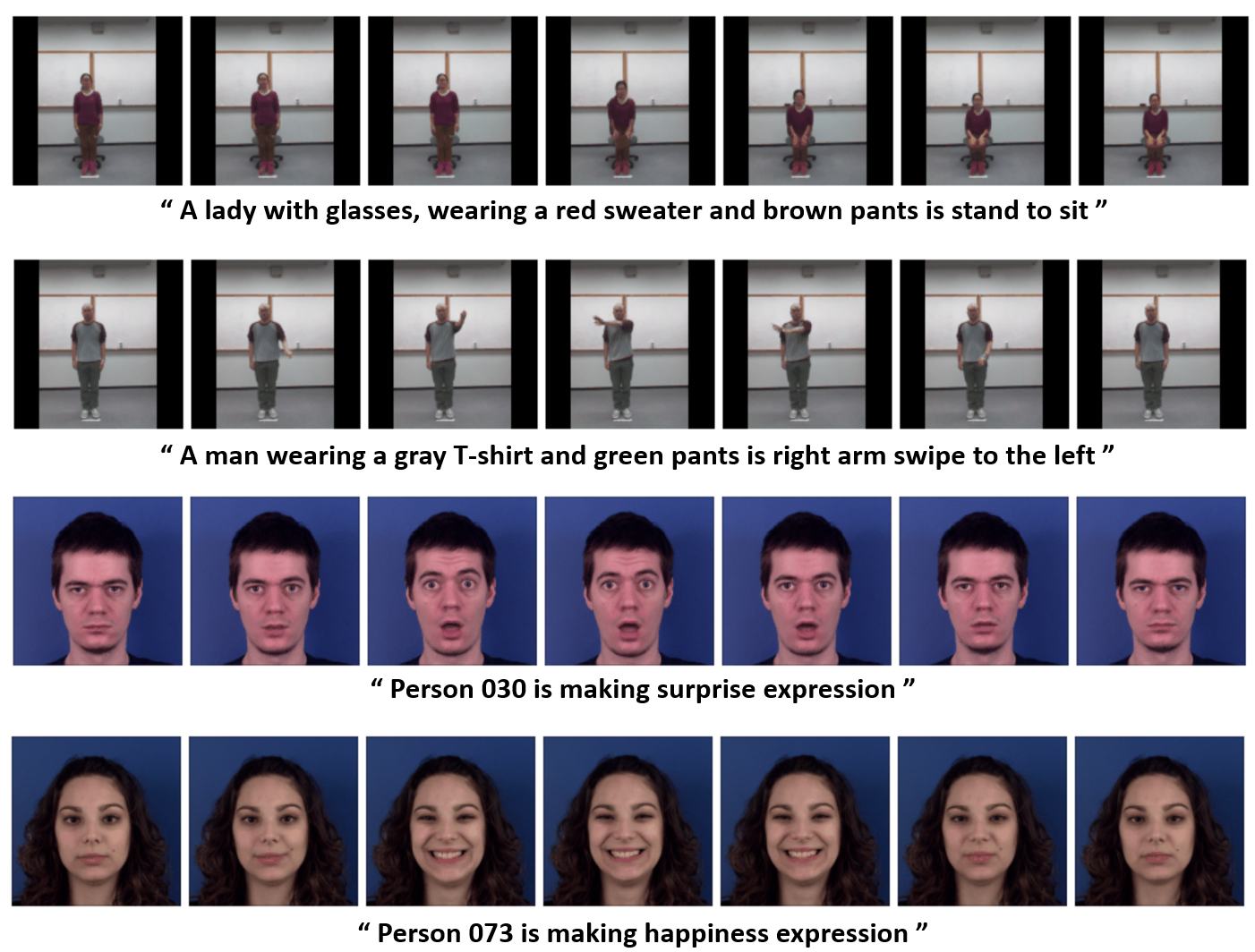}}
\vspace{-1mm}
\caption{Sector-Shaped Diffusion Model (S2DM) generates a video sample through a sector-shaped inverse diffusion area guided by two conditions. We also propose a two-stage generation strategy based on S2DM for high quality Text-to-Video generation task. More results can be viewd at~\href{https://s2dm.github.io/S2DM/}{https://s2dm.github.io/S2DM/}.}
\label{fig:result}
\end{center}
\vspace{-7mm}
\vskip -0.2in
\end{figure}

\section{Introduction}
\label{introduction}

Diffusion Probabilistic Models~\cite{sohl2015deep} (DPMs), which establish a forward diffusion process by adding noise to the original data and then learn the reverse denoising process to generate new samples from the random noise~\cite{song2020denoising,ho2020denoising}, have showcased its powerful generative capabilities in image generation tasks~\cite{dhariwal2021diffusion,nichol2021glide,saharia2022photorealistic,ramesh2022hierarchical}.
Recently, many researchers are exploring the adaptation of DPMs to video generation tasks~\cite{ho2022imagen,he2023latent}. 
Due to the scarcity of abundant and high-quality video training data, alongside the inherent complexity of the spatio-temporal relationships within videos, it is difficult to maintain the consistency and continuity across video frames. 
To address this challenge, many prior methods extended the pre-trained image generation models to video generation by introducing extra spatio-temporal attention modules, which were trained on video datasets to acquire temporal information~\cite{singer2022make,qi2023fatezero,blattmann2023align,xing2023simda,an2023latent}. 
As the inherent black-box nature of the attention module in these methods, the learning of temporal information in videos occurs implicitly, leading to a lack of interpretability and explicit control in modeling temporal features.
Some other methods opt to utilize extra information as additional guidance to enable the model to learn temporal features within the videos~\cite{yang2023rerender,ma2023follow}. 
%
Nevertheless, without specialized model designs, the incorporation of additional information is likely to introduce varying stochastic details during the video frame generation process, thereby compromising the consistency of the generated videos.

\begin{figure}[tb]
\vspace{-2mm}
\begin{center}
\centerline{\includegraphics[width=0.8\linewidth]{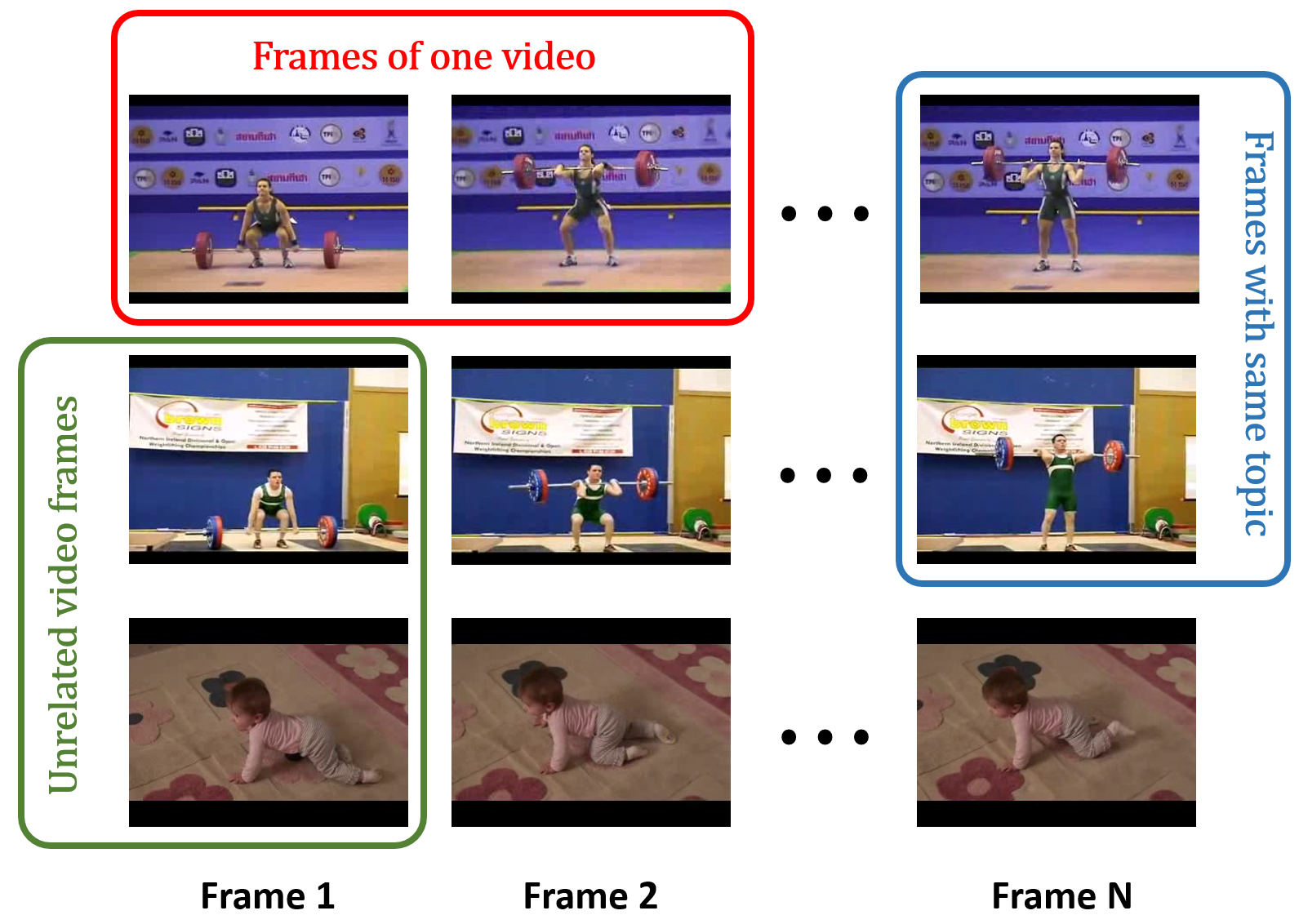}}
\vspace{-3mm}
\caption{Comparison between video frames under different scenarios. Each row represents a sequence of frames from a video: the above two rows depict "A person is doing weightlifting" while the last row illustrates "A little baby is crawling".}
\label{fig:video_features}
\end{center}
\vskip -0.2in
\vspace{-7mm}
\end{figure}


In this work, we posit that the content presentation of a video is primarily influenced by three key components: semantic-content features, temporal characteristics, and stochastic details.
As shown in Fig.~\ref{fig:video_features}, for any two unrelated videos, the characteristics of these three components should be different. 
For two distinct videos sharing the same thematic content, they may exhibit identical semantic-content features while differing in temporal characteristics and stochastic details. 
And within a video, the semantic-content features and stochastic details across frames should remain consistent, while temporal characteristics vary.
%
To generate high-quality videos, we identify three key challenges: ensuring semantic and content consistency throughout the video frames, retaining uniform stochastic characteristics from one frame to the next, and 
managing the frame-to-frame differences to accurately reflect temporal features.

\begin{figure}[ht]
\vspace{-1mm}
\begin{center}
\centerline{\includegraphics[width=1.0\linewidth]{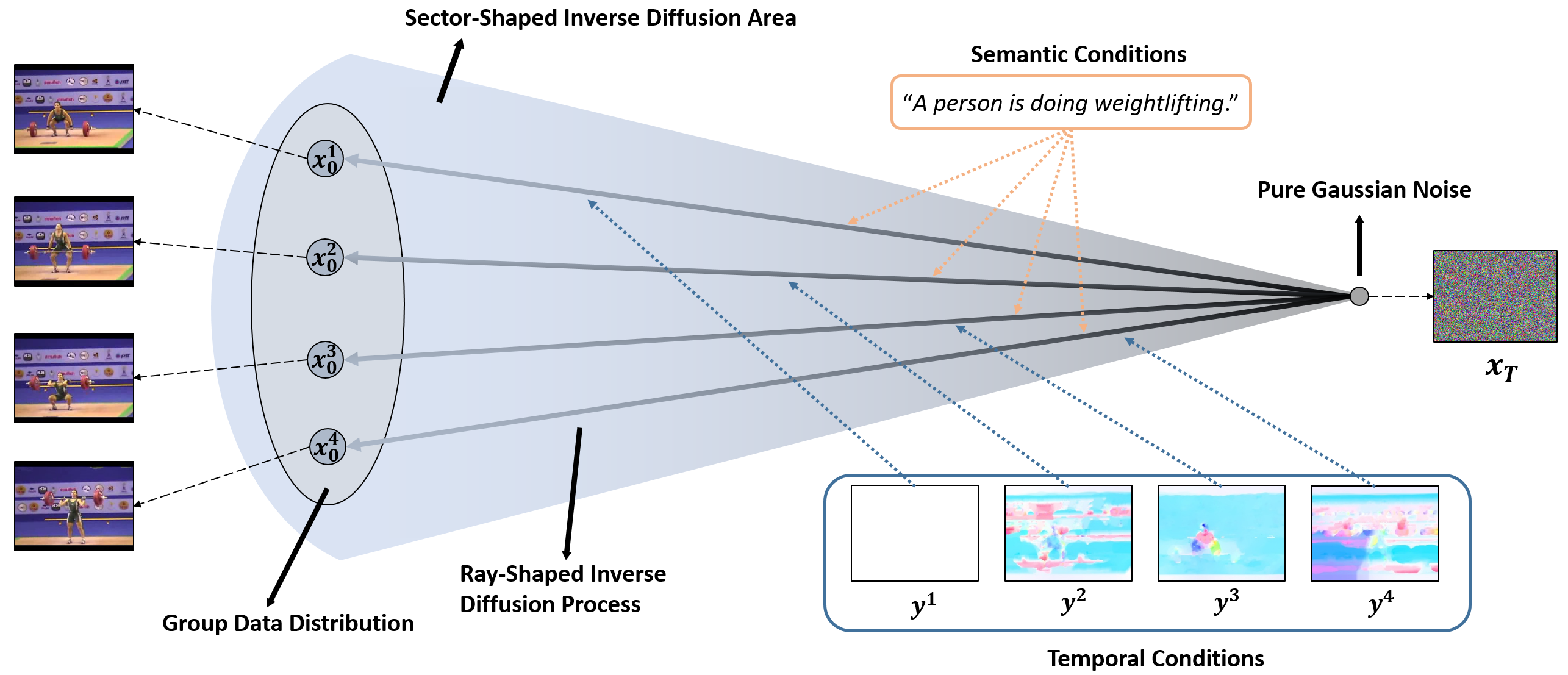}}
\vspace{-2mm}
\caption{Sector-Shaped Diffusion Model (S2DM): The sector-shaped inverse diffusion area is expanded by a set of ray-shaped inverse diffusion processes starting from the same initial noise point, guided by an identical semantic condition to maintain the content consistency and a set of temporal conditions to assign corresponding temporal features to each generated data point.}
\label{fig:sectorDM}
\end{center}
\vspace{-7mm}
\vskip -0.2in
\end{figure}

Following the assumption of Denoising Diffusion Probabilistic Models~\cite{sohl2015deep,ho2020denoising} (DDPMs) , if a group of data was continuously perturbed by the same noise, they would eventually approach to the same noise point.
By reversing this process, therefore, one can initiate it from a random noise point and generate a group of diverse data points with shared stochastic characteristics.
Based on this observation, we propose to model the generation of a group of intrinsically related data with shared stochastic features as a sector-shaped inverse diffusion area expanded by a set of ray-shaped inverse diffusion processes, starting from the same initial noise point, as illustrated in Fig.~\ref{fig:sectorDM}.
To infuse each data point within the group with uniform semantic content and distinct temporal features, we guide each reversing process with extra the same semantic condition and different temporal conditions across different data points respectively.

We apply this Sector-Shaped Diffusion Model (S2DM) in video generation tasks. 
We explore the use of text descriptions and optical flow as semantic and temporal conditions respectively.
%
%
%
%
We adopt the classifier-free~\cite{ho2022classifier} approach to incorporate guidance into the diffusion process.
Extensive experiments on a large number of datasets are conducted, and the results show that our methods achieves comparable results to current state-of-the-arts methods in generating video with high consistency.
We also designed an additional two-stage generation strategy to achieve the text-to-video generation task.

 




 Our contributions are summarized as follows:
 \begin{itemize}
     \item We propose that the key to address the the consistency challenge in video generation lies in maintaining semantic and content consistency across frames, ensuring uniform stochastic details between consecutive frames, and aligning the frame-to-frame differences with temporal characteristic.
     \item We propose a novel Sector-Shaped Diffusion Model (S2DM) to generate a group of intrinsically related data through a sector-shaped inverse diffusion area guided by two conditions at the same time.
     \item We apply our S2DM framework in video generation tasks and experiments results indicate the high generation quality of our methods.
 \end{itemize}

\vspace{-2mm}
\section{Related Work}
\label{related work}


%
Methods based on Denoising Diffusion Probabilistic Models (DDPMs)~\cite{ho2020denoising,sohl2015deep,song2020denoising} have broken the long-standing dominance of Generative Adversarial Networks (GANs)~\cite{goodfellow2014generative} in a variety of challenging generation tasks.
%
%
Prominent works such as DALL-E 2~\cite{ramesh2022hierarchical}, GLIDE~\cite{nichol2021glide} and Imagen~\cite{saharia2022photorealistic} leverage diffusion models for text-to-image (T2I) generation, employing the diffusion process in pixel space and guided by CLIP~\cite{radford2021learning} or T5~\cite{raffel2020exploring} through classifier-free approaches~\cite{ho2022classifier}. 
Latent Diffusion Model (LDM)~\cite{rombach2022high} uses an autoencoder~\cite{esser2021taming} to reduce pixel-level redundancy, significantly improving computational efficiency. For conditional image generation with flexible inputs, ControlNet~\cite{zhang2023adding}, T2I-Adapter~\cite{mou2023t2i} and Composer~\cite{huang2023composer} integrate a distinct spatial condition into the model, offering finer control over the generation process. 

When applying diffusion models to video generation tasks, one of the greatest challenge is to maintain the consistency and continuity across video frames. Among numerous attempts, VDM~\cite{ho2022video} pioneers in designing video diffusion models for Text-to-Video (T2V) generation, which extends the traditional image diffusion U-Net~\cite{ronneberger2015u} architecture to a 3D U-Net structure and adopts a joint training strategy of images and videos. 
Similarly, MagicVideo~\cite{zhou2022magicvideo}, based on LDM~\cite{rombach2022high}, introduces a frame-wise adapter to align the distributions of images and videos for effectively modeling temporal relationships. LVDM~\cite{he2023latent} also utilizes LDM as its backbone within a hierarchical framework based on 3D U-Net for latent space modeling. To reduce the reliance on video data collection, Make-A-Video~\cite{singer2022make} proposes a novel paradigm to learn visual-textual correlations from paired image-text data and video motion from unsupervised video data. Other works, for instance, SimDA~\cite{xing2023simda} and Video LDM~\cite{blattmann2023align} choose to introduce additional temporal blocks to learn temporal features in videos. While these approaches have demonstrated success in capturing temporal information within videos, the majority of the extra temporal modules are constructed in a black-box manner, resulting in a lack of interpretability for the overall model. 

Some studies aim for a more intuitive modeling of temporal information within videos through specialized model designs and theoretical insights. Latent-Shift~\cite{an2023latent} focuses on the lightweight temporal modeling. Taking inspiration from TSM~\cite{lin2019tsm}, it performs channel shifts between consecutive frames within convolution blocks to achieve temporal modeling. This strategy helps to guarantee consistency in generating frame motion, but also limits the generation diversity. Different from most approaches that denoise the frames independently, VideoFusion~\cite{luo2023videofusion} decomposes the diffusion process by employing a shared base noise and residual noise along the temporal axis for each frame. Similar to it, our model also makes the assumption of shared noise, too. However, we assume that the frames within one video are generated from one shared noise, and are perturbed with identical random noise during the training process, while they did not apply such shared noise assumption during the noise-adding process in training.

Other approaches choose to incorporate supplementary information to guide diffusion models, introducing additional guidance to facilitate the model in learning temporal features within the video. The common supplementary information are pose~\cite{hu2023animate,ma2023follow,xu2023magicanimate,wang2023disco} and optical flow~\cite{yang2023rerender,wang2023videocomposer,liang2023flowvid}. Animate Anyone~\cite{hu2023animate} encode the pose sequence with a lightweight Pose Guider and combine it with multi-frame noise to incorporate the posture information into the video generation process. Different from previous methods relying solely on optical flow, FlowVid~\cite{liang2023flowvid} encode the optical flow through warping from the initial frame, which serves as an additional reference in the diffusion model. These approaches are commonly employed in Video-to-Video (V2V) generation tasks. The supplementary information plays a crucial role in modeling the temporal characteristics, guiding the model to produce results enriched with robust temporal information, but the inclusion of extra information may introduce new stochastic details if the model is lacking of specialized design, which will impede consistent video generation.
\section{Sector-Shaped Diffusion Model with Shared Noise}

\subsection{Standard Diffusion Process}
A DDPM~\cite{ho2020denoising} consists of two $T$-step Markov chains: a forward process and a backward process. The forward process aims to continuously perturbs the data with the transition: 
\begin{equation}\label{eq: std forward}
    q(x_t|x_{t-1}) = \mathcal{N}(x_t;\sqrt{\alpha_t}x_{t-1},(1-\alpha_t)\mathbf{I}),
\end{equation}
where $t \in [ 0,T ]$ denotes the diffusion time step and $\alpha_t \in (0,1)$ is a hyperparameter determined before training with the constraint that
\begin{equation}\label{eq: alpha constraint}
    \lim\limits_{t \to \infty}\alpha_t = 0.
\end{equation}   
We incrementally introduce Gaussian noise to an image until eventually it becomes completely the same to pure Gaussian noise. The backward process aims to iteratively denoise the noise back to the original data state with the transition:
\begin{equation}\label{backward trasition}
    p_{\theta}(x_{t-1}|x_{t}) = \mathcal{N}(x_{t-1};\mu_{\theta}(x_t,t),\Sigma_{\theta}(x_t,t)),
\end{equation}
where the mean $\mu_{\theta}(x_t,t)$ and the variance $\Sigma_{\theta}(x_t,t)$ are parameterized by a deep neural network. With re-parameterization trick, one can obtain the learning objective in the $\epsilon$-prediction form:
\begin{equation}\label{eq: uncondtioned loss}
    \mathcal{L} = \mathbb{E}_{x_{0},t,\epsilon_t \sim\mathcal{N}(0,1)}||\epsilon_t - \epsilon_{\theta}(x_{t},t)||^2_2,
\end{equation}
which simplifies the problem to learn a network that predicts the noise at every time step $t$. If a condition $c$ is introduced during the diffusion process, the resulting learning objective can be represented as:
\begin{equation}\label{eq: conditioned loss}
    \mathcal{L} = \mathbb{E}_{x_{0},t,c,\epsilon_t \sim\mathcal{N}(0,1)}||\epsilon_t - \epsilon_{\theta}(x_{t},t,c)||^2_2.
\end{equation}

\subsection{Sector-Shaped Diffusion Process}\label{sector diffusion}

\subsubsection{Sector-Shaped Forward Diffusion Process}\label{sector-forward}
For a group of data $\mathbf{x} = \{ x^i_{0} | i = 1,2,\cdots, N \}$, we represent the noised result of $\mathbf{x}$ at time step $t$ as $\mathbf{x}_{t} = \{ x^i_{t} | i = 1,2,\cdots, N \}$. Following~\cref{eq: std forward} and the Markov property of forward diffusion process, the transition from $x_0^i$ to $x_t^i$ can be formulated as:
\begin{equation}\label{eq: forward transition}
    x_{t}^{i} = \sqrt{\hat{\alpha}_t}x_{0}^{i} + \sqrt{1 - \hat{\alpha}_t}\epsilon_{t}^{i},
\end{equation}
where $\epsilon_{t}^{i}\sim\mathcal{N}(0,1)$ is the random noise at diffusion step $t$ and $\hat{\alpha}_t = \prod\limits_{n=1}^t\alpha_n$ is a predefined hyperparameter. As one can see from~\cref{eq: alpha constraint}, $\hat{\alpha}_t$  will eventually converge to 0 as $t$ approaches $\infty$. Hence, if we kept the sampled noise to be the same across all data points in $\mathbf{x}$ at diffusion step $t$,
\begin{equation}
    \epsilon^i_t = \nu_t\sim\mathcal{N}(0,1),\quad i = 1,2,\cdots, N,
\end{equation}
then the noised result $x_t^i$ would be expressed as:
\begin{equation}\label{eq: forward transition shared noise}
    x_{t}^{i} = \sqrt{\hat{\alpha}_t}x_{0}^{i} + \sqrt{1 - \hat{\alpha}_t}\nu_{t}.
\end{equation}
With shared noise $\nu_t$, the noised result of $x_0^i$ will eventually converge to $\nu_T$ for all $x_0^i \in \mathbf{x}$ as $t$ approaches $\infty$. 

We abstract this process as a group of data points $\mathbf{x}$ within a distribution gradually converging towards an endpoint $\nu_T$, forming a sector-shaped region. The arc of this region comprises the original data points, with the center representing the final convergence endpoint. 

\subsubsection{Sector-Shaped Backward Diffusion Process}\label{sector-backward}
Considering the reverse process of~\cref{sector-forward}, we can initiate from a noise point and follow the backward paths within this sector-shaped diffusion region to generate a group of diverse data points, that share consistent stochastic details and semantic features among them, as shown in Fig.~\ref{fig:sectorDM}. We employ distinct temporal conditions to guide the generation of each data point, leading that although they start from the same initial point, the outcomes can exhibit different temporal characteristics. During the generation process of $x^i_0 \in \mathbf{x}$, we utilize temporal condition $y^i$ as additional guidance and according to~\cref{backward trasition}, we can obtain the backward transition as:
\begin{equation}\label{one cond backward trasition}
    p_{\theta}(x^i_{t-1}|x_{t}^i) = \mathcal{N}(x_{t-1}^i;\mu_{\theta}(x_t^i,t,y^i),\Sigma_{\theta}(x_t^i,t,y^i)).
\end{equation}

In addition, we also need another semantic condition $c$, which should stay the same for all $x^i_0 \in \mathbf{x}$, to guarantee that the final generated results share the same content and semantic characteristics. 
Now we can represent the backward transition as:
\begin{equation}\label{two cond backward trasition}
    p_{\theta}(x^i_{t-1}|x_{t}^i) = \mathcal{N}(x_{t-1}^i;\mu_{\theta}(x_t^i,t,y^i,c),\Sigma_{\theta}(x_t^i,t,y^i,c)).
\end{equation}
Similar to~\cref{eq: conditioned loss}, we can derive the $\epsilon$-prediction form learning objective of our sector-shaped diffusion process:
\begin{equation}\label{eq: loss two condition}
    \mathcal{L} = \mathbb{E}_{x_{0}^i,t,y^i,c,\nu_t \sim\mathcal{N}(0,1)}||\nu_t - \epsilon_{\theta}(x_{t}^i,t,y^i,c)||^2_2,
\end{equation}
where $\nu_t$ is the shared noise among all data points in $\mathbf{x}$ at diffusion step $t$. 


\subsection{Sector-Shaped Diffusion Process for Conditional Video Generation}\label{conditional generation}

\begin{figure}[ht]
\vskip 0.2in
	\centering
	\begin{minipage}{0.49\linewidth}
		\centering
		\includegraphics[width=\columnwidth]{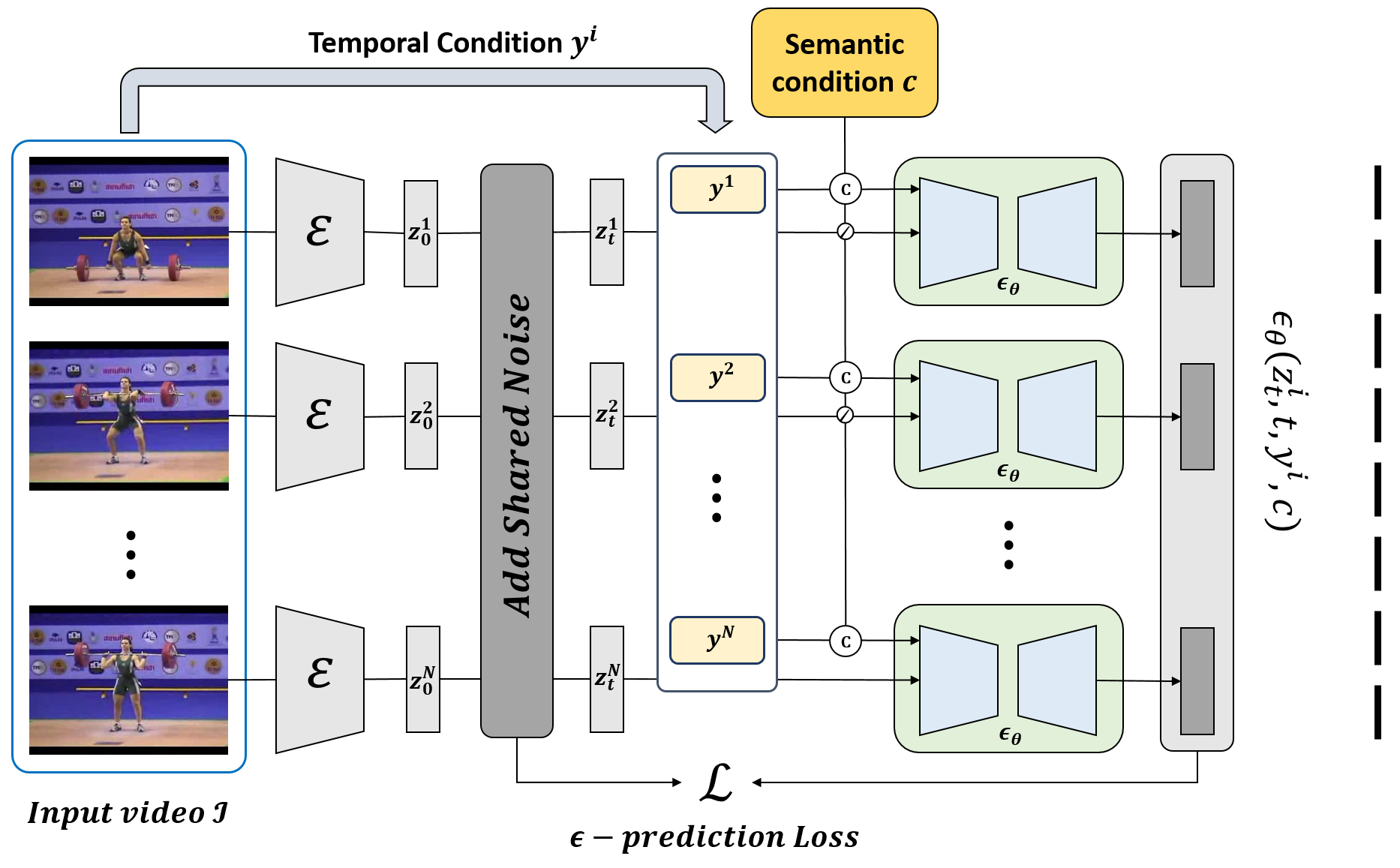}
	\end{minipage}
	\begin{minipage}{0.49\linewidth}
		\centering
		\includegraphics[width=\columnwidth]{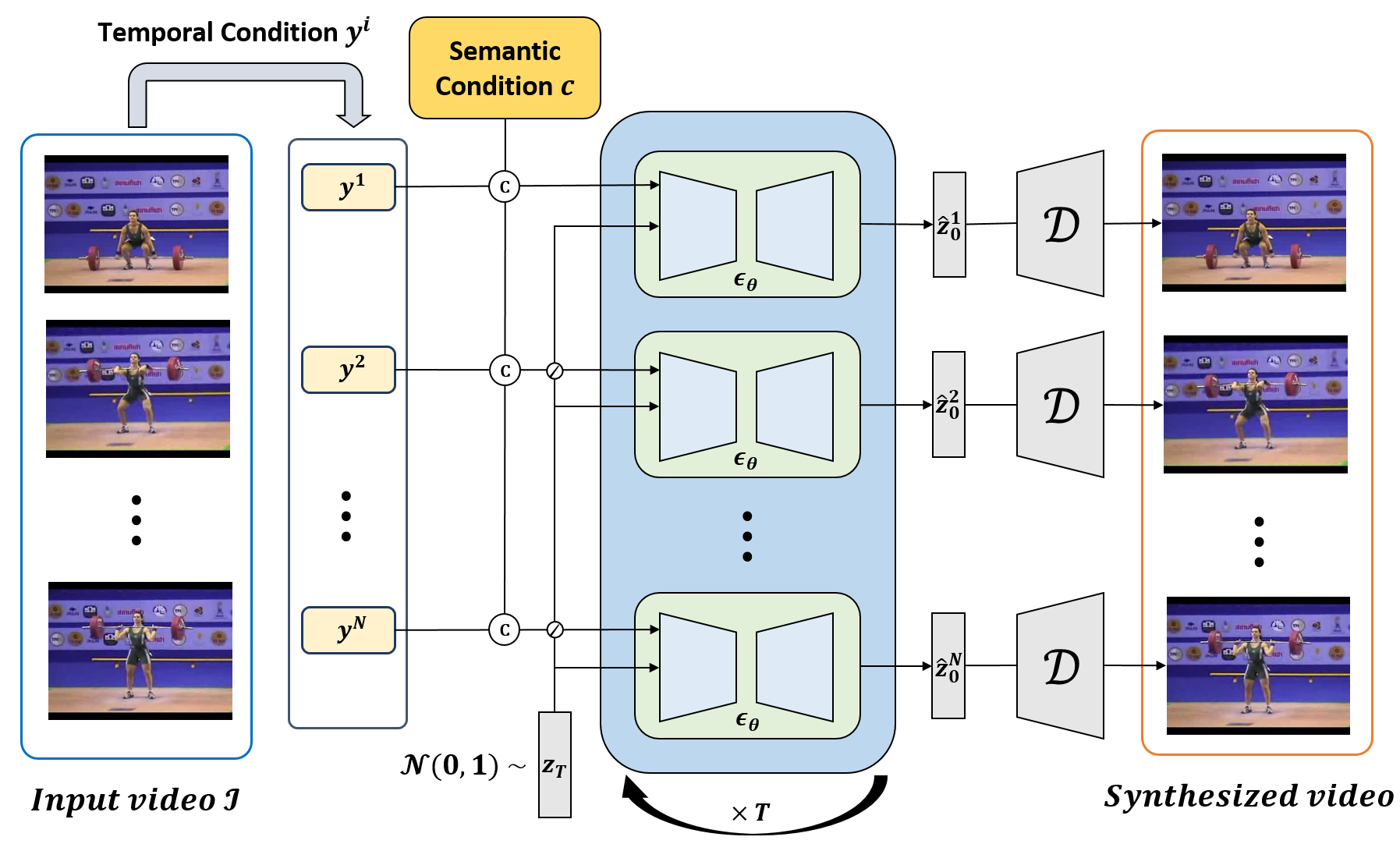}
	\end{minipage}
	\caption{\textbf{Training} (Left) and \textbf{Inference} (Right) stages of conditional video generation under Sector-Shaped Diffusion Model (S2DM) frame work (\cref{conditional generation}). During \textbf{training} stage, we perform shared-noise perturbation on the data from one video under the assumption of S2DM. Then we concatenate the extracted varying temporal conditions with the identical semantic condition to be the condition input of S2DM. As for \textbf{inference} stage, given temporal and semantic conditions, our S2DM iteratively generates various video frames based on the same initial random noise.}
	\label{common_caption}
\end{figure}

For conditional video generation tasks, the objective is to generate a video clip with $N$ frames $\mathcal{I} = \{I^1, \cdots, I^N\}$ conditioned on a semantic condition $\mathbf{c}$ and a set of temporal conditions $\mathbf{y} = \{y^1,\cdots, y^N\}$.
The semantic condition $\mathbf{c}$ typically describes the content of a video, such as text description or image context,
while temporal condition $\mathbf{y}$ commonly incorporates partial information about the video, including depth maps, optical flow and poses. They can be provided by the user or extracted from a reference video $\hat{\mathcal{I}} = \{\hat{I}^1, \cdots, \hat{I}^N\}$. 
In this work, 
we explore the use of optical flow as temporal conditions for conditional video generation.

\subsubsection{Optical flow-guided Video Generation}\label{flow guided}
We build our model based on a pre-trained T2I generation LDM~\cite{rombach2022high} that conduct diffusion process in the latent space. Specifically, with an encoder $\mathcal{E}$ to compress the image $I \in \mathbb{R}^{H \times W  \times 3}$ to a latent variable $z \in \mathbb{R}^{H/8 \times W/8  \times 4}$ and a text prompt $\tau$, LDM learns a U-Net~\cite{ronneberger2015u} parameterized by $\theta$ to estimate the noise added at diffusion step $t$ with loss function:
\begin{equation}\label{eq: LDM loss}
    \mathcal{L}_{LDM} = \mathbb{E}_{z_{0},t,\tau,\epsilon_t \sim\mathcal{N}(0,1)}||\epsilon_t - \epsilon_{\theta}(z_{t},t,\tau)||^2_2.
\end{equation}

For a video clip $\mathcal{I} = \{I^1, \cdots, I^N\}$, we calculate the backward flow $f^i$ between the first frame $I^1$ and the other frames $I^i$. The optical flow quantifies the motion between two frames in both horizontal and vertical directions, so it is expected to be zero for identical frames. 
%
%


We extend the text-to-image generation LDM to a per-frame video generation sector-shaped diffusion model by introducing the optical flow $\mathcal{F} = \{f^1, \cdots, f^N\}$ as temporal condition. 
We concatenate the condition $f^i$ with the CLIP~\cite{radford2021learning} embedding of text prompt $\tau$ along the channel dimension, and incorporate them into the diffusion model through cross attention.
During training, we follow classifier-free guidance~\cite{ho2022classifier} strategy to replace the condition $f^i$ and $\tau$ by a null label $\emptyset$ with a fixed probability respectively.

\begin{figure}[th]
\begin{center}
\centerline{\includegraphics[width=1.1\columnwidth]{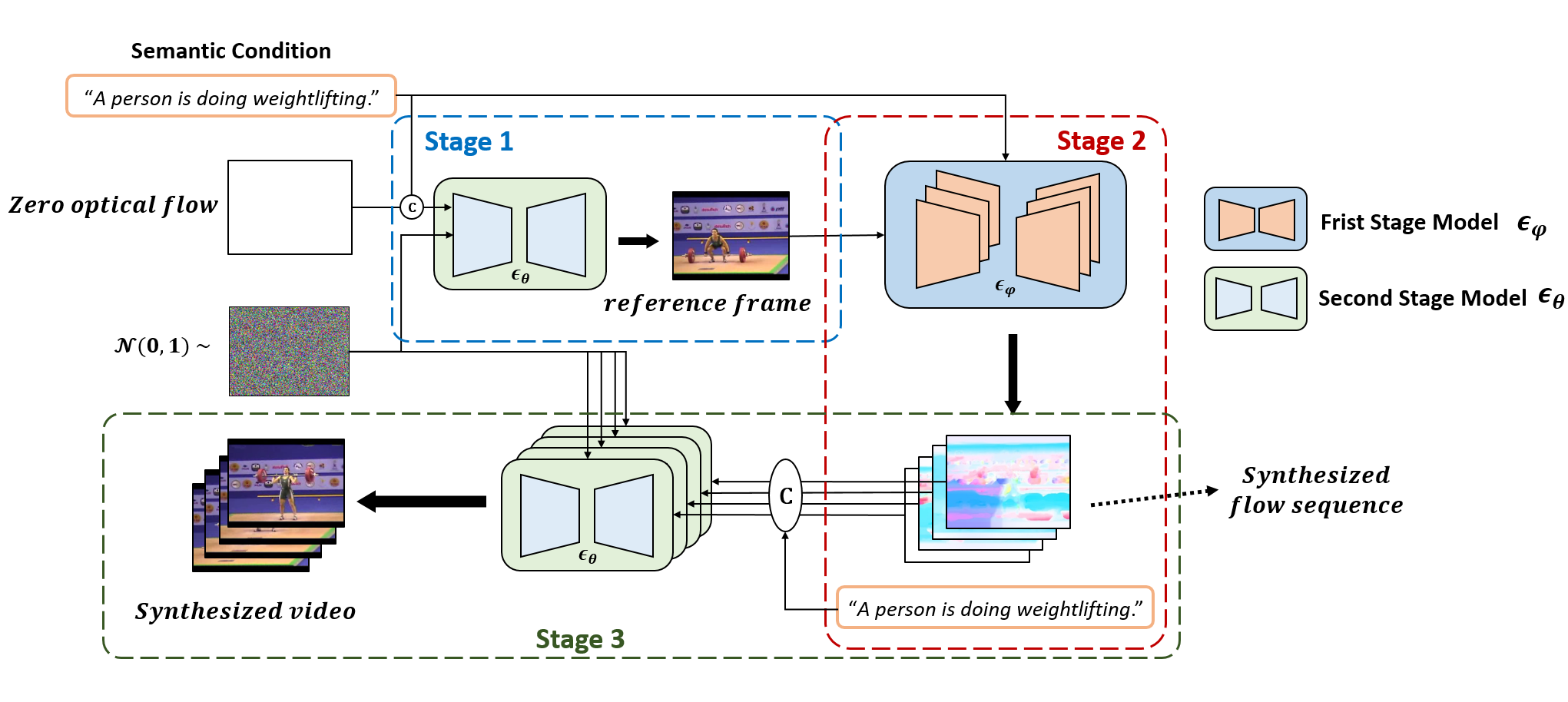}}
\caption{Inference process of Two-stage text-to-video generation pipeline (\cref{text-to-video generation}). \textbf{Firstly}, we employ the second-stage model $\epsilon_{\theta}$ to generate a reference frame $I_{ref}$, given text prompt $\tau$ and zero optical flow $f^1$. \textbf{Secondly}, we generate an optical flow sequence $\hat{\mathcal{F}}$ via the first stage model $\epsilon_{\phi}$ conditioned on text prompt $\tau$ and reference frame $I_{ref}$. \textbf{Thirdly}, we generate a sequence of video frames by the second-stage model $\epsilon_{\theta}$ conditioned on text prompt $\tau$ and synthesized flow sequence $\hat{\mathcal{F}}$, with the identical initial random noise sampled in first step. }
\label{fig:text-to-video}
\end{center}
\vskip -0.4in
\end{figure}

\subsection{Two-stage generation strategy for Text-to-Video Generation}\label{text-to-video generation}

The proposed S2DM framework is naturally suitable for conditional video generation tasks, but when it comes to Text-to-Video (T2V) generation tasks, we only have access to semantic condition$--$text, necessitating the generation of temporal conditions to guide the video generation process. Inspired by LFDM~\cite{ni2023conditional} and Diffusion AutoEncoders~\cite{preechakul2022diffusion}, we take a two-stage generation strategy, utilizing another DDPM as a first-stage model to generate temporal conditions used in our S2DM, following which is the second-stage model for generating video frames based on semantic and temporal conditions. 

The first-stage model consists of a diffusion model based on 3D U-Net~\cite{ho2022video}, denoted as $\epsilon_{\phi}$, which is designed to learn temporal information within the video. The model $\epsilon_{\phi}$ takes a reference image $I_{ref}$ and text prompt $\tau$ as conditions to generate a sequence of temporal features, such as a flow sequence, which will serve as the temporal conditions in the second-stage model for conditional video generation.
This two-stage video generation strategy decouples the generation of semantic-content features from temporal features, ensuring semantic content consistency across video frames and aligning frame differences with temporal features.

 In this work, we choose the optical flow-guided video generation model introduced in~\cref{flow guided} to serve as the second-stage model. Our inference process comprises three stages as follow:
 \begin{itemize}
     \item\label{stage 1} We first adopt the second-stage model condition on a text prompt $\tau$ and a zero optical flow $f^1$ to generate a reference frame $I_{ref}$, usually the first frame.
     \item Then we use the reference frame $I_{ref}$ along with the text prompt $\tau$ to generate an optical flow sequence $\hat{\mathcal{F}}$ by the second-stage model.
     \item  Finally we can generate the video frames using second-stage model condition on the text prompt $\tau$, optical flow sequence $\hat{\mathcal{F}}$ and the same noise we sampled in stage one.
 \end{itemize}

\section{Experiments}


%
%

\subsection{Experimental Setup}
\textbf{Datasets.} For optical flow-guided video generation (\cref{flow guided}), we train and test our method on two datasets: MHAD~\cite{chen2015utd} and MUG~\cite{aifanti2010mug} for quantitative evaluation. 
MHAD human action data set contains 861 videos of 8 human subjects performing 27 types of actions and MUG facial expression dataset contains 1,009 videos of 52 different human subjects performing 7 different expressions. We randomly select 36 out of 364 individual-expression combinations from MUG dataset and 21 out of 216 individual-action combinations from MHAD dataset as testing dataset respectively, which produce 78 testing videos and 931 training videos on MUG dataset; 84 testing videos and 777 training videos on MHAD dataset. 
In our finalized test split, the combinations of individuals and their specific actions/expressions are not found within the training set.
%
For each video, we craft a textual description, detailing the characteristics of the characters and the actions/expressions depicted in the video, to be the semantic condition for our S2DM. We also conduct two-stage Text-to-Video generation (\cref{text-to-video generation}) experiments on these two datasets.


We resize all videos to 128$\times$128 resolution for our model. And for each training video, we evenly divide all video frames into 16 intervals and randomly sample one frame from each interval, thereby obtaining a total of 16 video frames.

\textbf{Evaluation Metrics.} We quantitatively evaluate our model by Fréchet Video Distance (FVD)~\cite{unterthiner2018towards} and Kernel Video Distance (KVD)~\cite{unterthiner2018towards}. In our evaluation, we randomly generate 2048 videos conditioned on semantic and temporal conditions from the test dataset for all models, ensuring an accurate estimation of the distribution for computing the FVD score and KVD score. 

\textbf{Implementation Details.}\label{implementation details} We build our model based on the publicly available Stable Diffusion (SD)~\cite{ramesh2022hierarchical} checkpoints SD 1.5~\cite{rombach2022high}. Since SD 1.5 is trained on 512$\times$512 resolution images, directly applying it to the smaller-sized videos would result in image quality degradation, therefore we fine-tune the encoder and decoder on our experimental datasets with uncorrelated images randomly selected from different videos. To implement the shared noise assumption of the sector-diffusion model in training process, we use 4 videos as a training minibatch, and perturb the frames within one video with the same random noise. We train our model on a 4-A40-40G node, with the batch size of 1 for each GPU, and employ Adam optimizer~\cite{kingma2014adam} with a learning rate of 2e-5 to during training process.

During generation, we first sample the temporal conditions and semantic condition required for S2DM to generate videos (semantic condition only for Text-to-Video generation). With the sampled conditions, we perform a 20-steps DDIM~\cite{song2020denoising} sampling process on S2DM to generate each frame of the video, following classifier-free guidance~\cite{ho2022classifier} with a scale of 7.5. We generate 16 frames for a video in this manner and use them to test the performance of our model.

\begin{table}[tb]
\caption{Quantitative comparisons on MHAD and MUG. $\downarrow$ indicates the lower the better.}\label{table: MHAD and MUG result}
\setlength{\tabcolsep}{5pt} 
\centering
\begin{tabular}{cccccccc}
\toprule
\textbf{Method} &
\multicolumn{2}{|c|}{\textbf{Conditions}} &
\multicolumn{2}{|c|}{\textbf{MHAD}} &
\multicolumn{2}{|c}{\textbf{MUG}} &
\\
\midrule
 \phantom{\textbf{Method}} & 
{Text} & 
{Optical flow} &
{FVD $\downarrow$} &
{KVD $\downarrow$} & 
{FVD $\downarrow$} &
{KVD $\downarrow$} \\
\midrule
 LFDM~\cite{ni2023conditional} & $\surd$ & $\times$ &  148.58 &  9.95 & 105.04 & 1.61\\
 VDM~\cite{ho2022video}  & $\surd$ & $\times$ & 642.48 & 83.25 & 367.65 & 28.33\\
 LVDM~\cite{he2023latent} & $\times$ & $\times$ & 257.94 & 62.07 & 153.11 & 13.75\\
\midrule
 Ours (w/. optical flow) & $\surd$ & $\surd$ & $\mathbf{99.02}$ & $\mathbf{2.79}$ & $\mathbf{95.98}$ & $\mathbf{1.37}$\\
 Ours (w/o optical flow) & $\surd$ & $\times$ & 128.26 & 4.66 & 111.74 & 1.81\\
\bottomrule
\end{tabular}
\end{table}

\subsection{Optical Flow-Guided Video Generation}\label{exp:optical}
We compare our model with three baseline models: LFDM~\cite{ni2023conditional}, VDM~\cite{ho2022video} and LVDM~\cite{he2023latent}. 
%
LFDM is a conditional Image-to-Video (cI2V) generation model. It consists of a 3D U-Net-based diffusion model, which can generate a sequence of optical flow data given a class condition and a reference image. Then it warps the reference image in the latent space with the generated optical flow sequence, which are fed into another generator for synthesising a video.
%
As in LFDM, we obtain optical flow data for training through RegionMM~\cite{siarohin2021motion}, which provides both optical flow $\hat{f}^i \in \mathbb{R}^{h \times w \times 2}$ and occlusion map $\hat{o}^i \in \mathbb{R}^{h \times w \times 1}$ between $I^1$ and $I^i$ in the latent space. The occlusion map $\hat{o}^i$, values from 0 to 1, aids optical flow in image warping by conveying information about occlusions between two frames. We concatenate them as one vector $f^i \in \mathbb{R}^{h \times w \times 3}$ to be the flow condition used in our method. We reproduced VDM and LVDM based on the papers and trained them on the dataset. LVDM is an unconditional generation model, while VDM takes textual information as condition. As for LFDM, we use the pretrained models on MHAD and MUG dataset from their work~\cite{ni2023conditional}, conditioned on the text prompt and reference image.

As one can see in Table~\ref{table: MHAD and MUG result}, the results show that our S2DM achieves optimal performance when utilizing both text prompts and optical flow information (ours w/. optical flow) on both the MHAD and MUG datasets.
Notably, the strongest baseline, LFDM, relies on a reference image to generate a video, which may introduce potential bias in FVD and KVD computation, as information from the test set is inadvertently exposed to the model through the reference image.
However, our S2DM still demonstrates improvements over it, achieving enhancements of 33.36\% in FVD and a noteworthy increase of 71.96\% in KVD scores on MHAD dataset, and showcasing a improvement of 8.63\% in FVD and 14.91\% in KVD scores on MUG dataset. 
We believe this is due to the fact that, in our approach, optical flow is involved in the video generation process in a conditional manner, differing from the warp method employed in LFDM. This help us to avoid certain issues that may arise from rule-based methods, such as potential image distortions.

We visualize some of generated results on MHAD and MUG dataset in Fig~\ref{fig:result}. As evident from these results, S2DM generates video frames that align with the text prompts. More importantly, stochastic details such as the character's appearance (including the color of its clothes or hair), lighting conditions, and the background of each frame within a video are consistent. More results are displayed at~\href{https://s2dm.github.io/S2DM/}{https://s2dm.github.io/S2DM/}.



\begin{figure}[ht]
    \centering
    \begin{subfigure}[b]{1.0\textwidth}
        \centering
        \includegraphics[width=\textwidth]{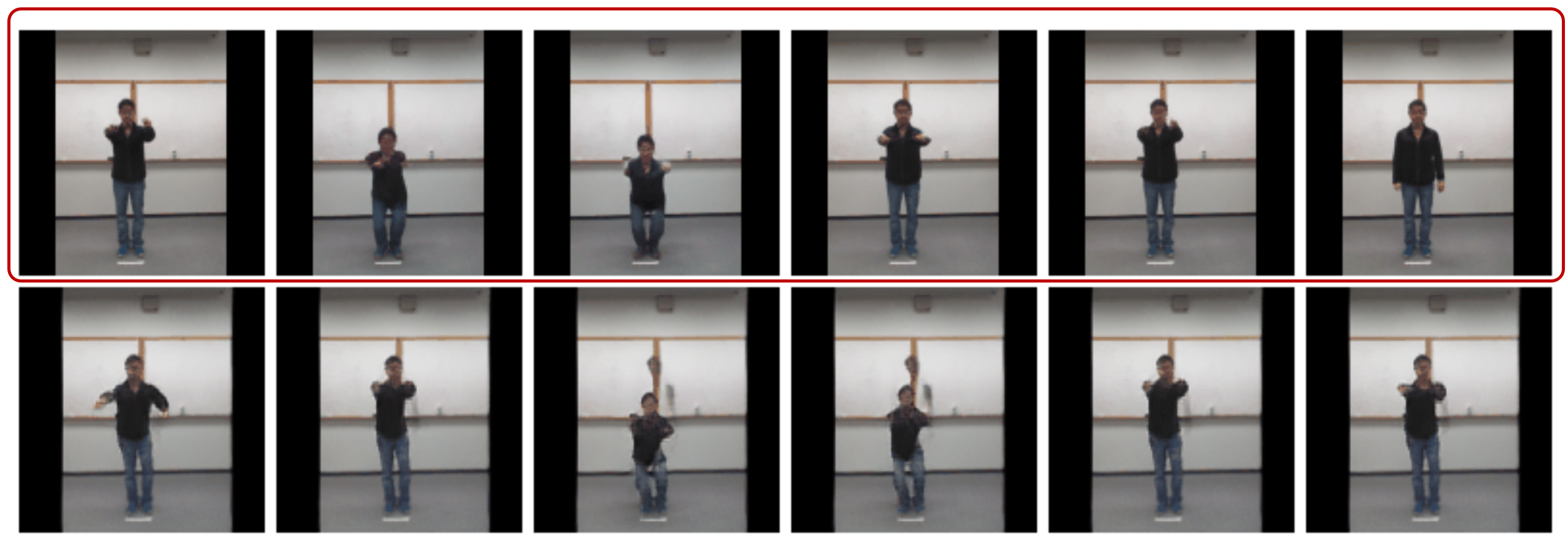}
        \caption{" A man with glasses, wearing a black coat and blue jeans is squat " (MHAD)}
        \label{fig:sub1}
    \end{subfigure}
    \quad
    \begin{subfigure}[b]{1.0\textwidth}
        \centering
        \includegraphics[width=\textwidth]{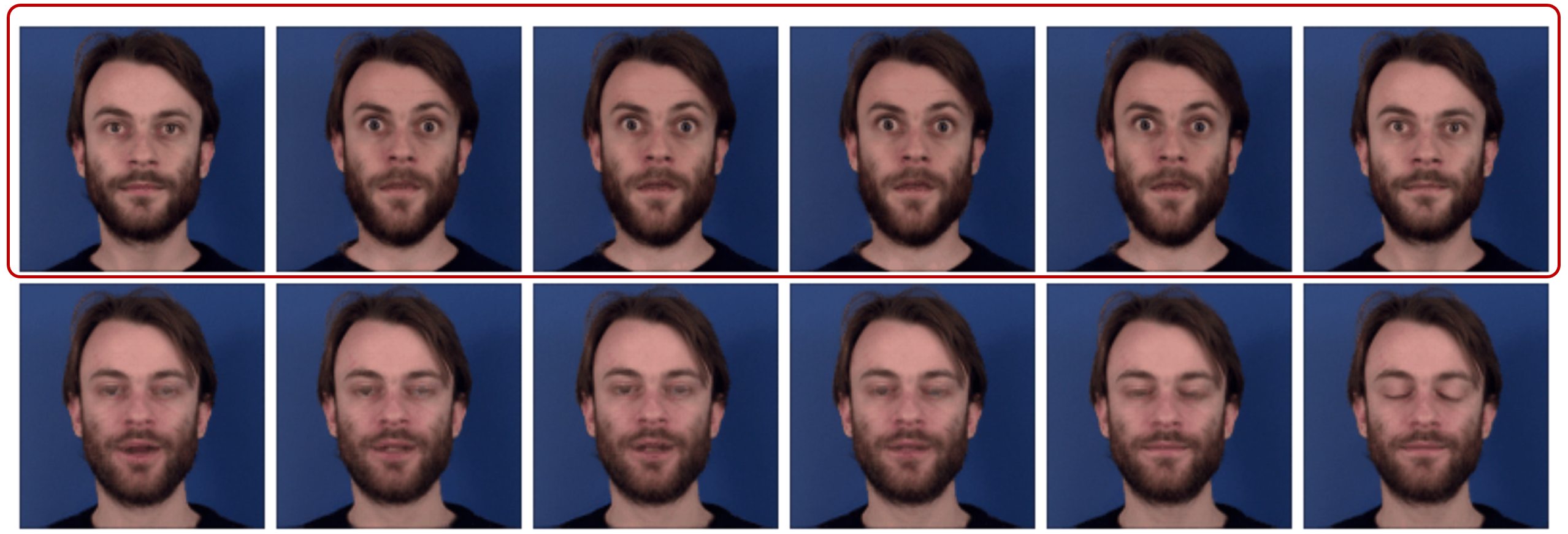}
        \caption{" Person 020 is making fear expression " (MUG)}
        \label{fig:sub2}
    \end{subfigure}
    \caption{Qualitative comparison with LFDM~\cite{ni2023conditional} on two datasets for two-stage Text-to-Video generation. The first row shows videos generated by our two-stage model based on S2DM, while the second display videos generated by LFDM conditioned both text prompt and a reference image.}
    \vspace{-3mm}
    \label{fig:compare}
\end{figure}

\subsection{Two-stage Text-to-Video Generation}\label{exp:t2v}
Following the two-stage generation strategy introduced in section~\ref{text-to-video generation}, we also conduct a Text-to-Video generation experiment on MHAD and MUG dataset. We apply the flow generation model in LFDM~\cite{ni2023conditional} to be the first-stage model in our method.  


As shown in Table~\ref{table: MHAD and MUG result}, our two-stage T2V generation method (ours w/o optical flow) surpasses most baseline models, except for LFDM on MUG dataset. It is worth noting that the optical flow condition used for our second stage model is derived from the flow generated by the LFDM model, which is different from the optical flow condition used during model training. This constitutes a zero-shot test for the optical flow condition, potentially influencing the final generated results to some extent. On MHAD dataset, our method outperforms the best baseline model LFDM by 13.68\% and 53.18\% respectively in FVD score and KVD score. On MUG dataset, our two-stage generation method slightly trails behind LFDM, yet still significantly outperforms other models.
We attribute this to the varying nature of the datasets. In MUG dataset, which primarily focuses on facial expressions with intricate details, a good reference image significantly boosts performance. Conversely, MHAD dataset deals with action sequences with less detail, where video quality is more reliant on overall motion patterns.

Some qualitative comparisons between the results of ours and LFDM are depicted in Fig.~\ref{fig:compare}. On MHAD dataset, our model demonstrates a clear advantage over LFDM in terms of both video consistency and generation quality. The results of LFDM often presents distortions and noticeable deformations. Although the disparity between the generated results of our models on MUG dataset is comparatively smaller, distortions are still apparent in certain details like the mouth and eyes. Additionally, LFDM-generated videos on MUG dataset exhibit significant jitter issues, which are less prevalent in videos generated by our model.


\subsection{Ablation Study}


\begin{table}[tb]
\setlength{\tabcolsep}{5pt} 
\caption{Ablation study of different noise schedule.}
\vspace{-3mm}
\centering
\begin{tabular}{cccccc}
\toprule
\multicolumn{2}{c|}{\textbf{Noise Schedule}} &
\multicolumn{2}{|c}{\textbf{MHAD}} &
\multicolumn{2}{|c}{\textbf{MUG}}\\
\midrule
Training & Sampling & FVD $\downarrow$ & KVD $\downarrow$ & FVD $\downarrow$ & KVD $\downarrow$ \\
\midrule
 Shared noise & Shared noise &  $\mathbf{99.02}$ &  $\mathbf{2.79}$ & $\mathbf{95.98}$ & $\mathbf{1.37}$ \\
 Shared noise & Non-shared noise & 102.81 & 3.28 & 124.98 & 2.35 \\
 Non-share noise & Non-shared noise & 114.28 &  4.47 & 130.07 & 2.44\\
\bottomrule
\vspace{-4mm}
\end{tabular}
\label{table: ab noise schedule}
\end{table}

\noindent\textbf{Shared Noise Assumption}\quad A critical question we want to answer is how effective the proposed shared noise diffusion framework is? To validate that, we study three different assumptions regarding noise schedule during the video training and sampling processes: (I) adding shared noise in training stage and start from the same noise at sampling stage (II) shared noise in training stage while start from different random noises at sampling stage (III) non-shared noise for both training and sampling stage, which is the same as the standard diffusion process. 

We conduct the experiment based on optical flow-conditioned video generation introduced in~\cref{flow guided} on MHAD and MUG dataset. 
We use fully-trained S2DM to generate videos for noise schedule (I) and (II). For noise schedule (III), we retrain the model on the dataset without shared noise strategy and generate videos via non-shared noise sampling.
As shown in Table.~\ref{table: ab noise schedule}, the results under the shared noise schedule significantly outperform the other two noise schedules.
Compared to non-shared noise schedule (III), the result of shared noise schedule (I) leads by 13.35\% and 37.58\% respectively in FVD score and KVD score on MHAD dataset, and by 26.21\% and 43.85\% respectively on MUG dataset.
This is primarily due to the shared noise assumption in our S2DM framework, which effectively maintains consistency between content features and stochastic details across video frames, thereby elevating the overall quality of the generated videos. The dynamic comparison results can be viewed at~\href{https://s2dm.github.io/S2DM/}{https://s2dm.github.io/S2DM/}.
\section{Limitations and Future Work}
While our proposed Sector-Shaped Diffusion Model (S2DM) has achieved certain effectiveness in optical-flow guided video generation tasks and text-conditioned video generation tasks, it still has some limitations. Firstly, our current method of incorporating semantic condition and temporal condition is relatively straightforward, lacking the ability to effectively separate and decouple these two conditions, which may hinder the quality of generated videos. Moving forward, we aim to explore introducing these two conditions separately at different stages of diffusion process to enhance control over the generation process. 
Secondly, we only explore optical flow as temporal conditions in this work. To further demonstrate the generality of conditions in our S2DM framework, we plan to use different conditions as semantic condition and temporal conditions in our future work, for example, utilizing the posture information as the temporal condition and a reference image as the semantic condition. 
Furthermore, we may over simplify how a video content being influenced by its semantic, temporal and stochastic features in our work. In real world, a long and content-rich video may have more complex structures. However, our work could still contribute to research on consistent video generation and S2DM can be treated as a basic building block which is able to preserve stochastic details within a sequence of video frames. In our future work, following the idea of two-stage generation strategy, we will plan to compose multiple S2DMs, where each is guided by different conditions, to generate more complex and longer videos. 
\section{Conclusion}
In this work, we propose a novel Sector-Shaped Diffusion Model (S2DM) to generate a group of intrinsically related data sharing the same semantic and stochastic features while varying on temporal features. In contrast to most diffusion-based methods, we model the generation process of data as a sector-shaped diffusion region formed by a set of ray-shaped reverse diffusion processes initialize from the same noise point. We apply S2DM to video generation tasks, utilizing optical flow as temporal conditions. Experimental results demonstrate that our model performs well in maintaining the consistency and coherence of generated videos.

%
%
\bibliographystyle{splncs04}
\bibliography{main}
\end{sloppypar}
\end{document}